\documentclass[11pt,twocolumn]{article}
\usepackage[english]{babel}
\usepackage{amsmath,amsthm}
\usepackage{amsfonts}
\usepackage{graphicx}
\usepackage{url}
\usepackage{amssymb}
\usepackage{multicol}
\usepackage{wrapfig}
\usepackage{subfig}
\usepackage{url}

\usepackage{vmargin}
\setpapersize{A4}
\setmargins{1.5cm}{1.5cm}{18cm}{22,7cm}{1cm}{1cm}{1cm}{2cm}

\title{\textbf{From the decoding of cortical activities\\ to the control of a JACO robotic arm:\\ a whole processing chain}}%
\author{Laurent Bougrain$^{1,2}$, Olivier Rochel$^{2}$, Octave Boussaton$^{2}$ and Lionel Havet$^{2}$}%
\date{ $^1$ Universit{\'e} de Lorraine, LORIA, UMR 7503, Vandoeuvre-l{\`e}s-Nancy, F-54506, France\\$^2$ Inria, Villers-l{\`e}s-Nancy, F-54600, France}%

\begin{document}

\begin{multicols}{1}
\vspace{-5cm}
\maketitle


\end{multicols}



\begin{abstract}
This paper presents a complete processing chain for decoding intracranial data recorded in the cortex of a monkey and replicates the associated movements on a JACO robotic arm by Kinova. We developed specific modules inside the OpenViBE platform in order to build a Brain-Machine Interface able to read the data, compute the position of the robotic finger and send this position to the robotic arm. More precisely, two client/server protocols have been tested to transfer the finger positions: VRPN and a light protocol based on TCP/IP sockets. According to the requested finger position, the server calls the associated functions of an API by Kinova to move the fingers properly. Finally, we monitor the gap between the requested and actual fingers positions. This chain can be generalized to any movement of the arm or wrist.

\end{abstract}

\vspace{0.5cm}

\thispagestyle{empty}


\section{Introduction}

  \begin{figure*}[htbp]
    \begin{center}
     \includegraphics[width=150mm]{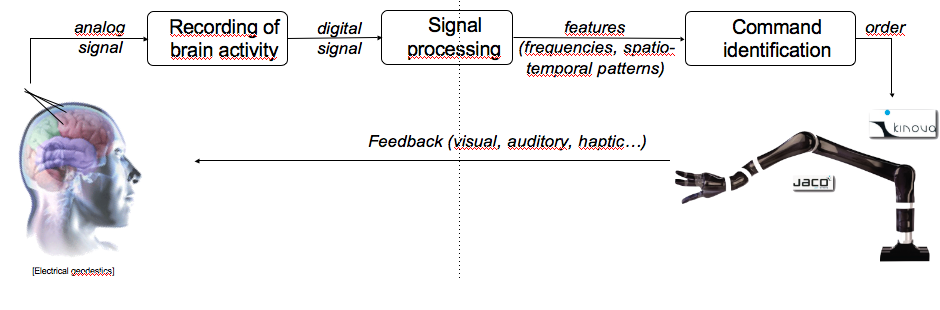}
      \caption{Brain-Computer Interface loop : from brain imaging, feature extraction and classification to feedback.}
      \label{fig:bciloop}
    \end{center}
  \end{figure*}
  
Brain-Computer interfaces (BCI) \cite{Wolpaw2002} interpret brain activity to produce commands on a computer or other devices like robotic arms. A BCI therefore allows its user, and especially a severely impaired person, to interact with their environment only using their brain activity.

The main applications of BCIs are the assistance to people with motor disabilities (especially those completely paralyzed suffering from locked-in syndrome), multimedia (eg video games), and more generally the interaction with any automated system (robotics, home automation, etc.). A BCI can be described as a closed loop system (Fig. \ref{fig:bciloop}) consisting of six main steps:
\begin{enumerate}
\vspace{-0.2cm}\item recording of brain activity (electroencephalography in particular);
\vspace{-0.3cm}\item signal processing (to remove artefacts, apply a band-pass filter);
\vspace{-0.3cm}\item feature extraction (to amplify and recover useful information);
\vspace{-0.3cm}\item classification (to identify the order);
\vspace{-0.3cm}\item translation of the order into a command;
\vspace{-0.3cm}\item feedback (to provide information on the outcome of the order and/or brain activity).
\end{enumerate}

More specifically, impressive results such as \cite{Santucci2003} and \cite{Velliste2008} exist in the neuroscience literature where the cortical activity of a monkey is used to control a complete robotic arm. More and more studies are dedicated to better understand motor control. It is now possible to decode from intracranial data the direction of a movement, the finger flexion or the muscular effort.

Following our long time study of the prediction of the finger flexion from ECoG signals \cite{Liang2009}  and noninvasive BCI, we are now concerned, as part of a regional initiative\footnote{PRST MISN/Situated Informatics (\url{http://infositu.loria.fr})}, with the correct control of a robotic arm in a Brain-Machine Interface. Hence the necessity for a complete processing chain, we developed a solution to control a JACO robotic arm by Kinova from brain signals using the OpenViBE software\footnote{\url{http://openvibe.inria.fr}}. In section \ref{material}, we will introduce the experimental setup, the recorded data, OpenVIBE and the JACO robotic arm. Section  \ref{method} will present our contribution in terms of new OpenViBE modules, client/server applications and interface for controlling the JACO robotic arm. Then, in the conclusion, we will share our accumulated experience.

\section{Material}
\begin{figure}[b!]
    \begin{center}
     \includegraphics[width=87mm]{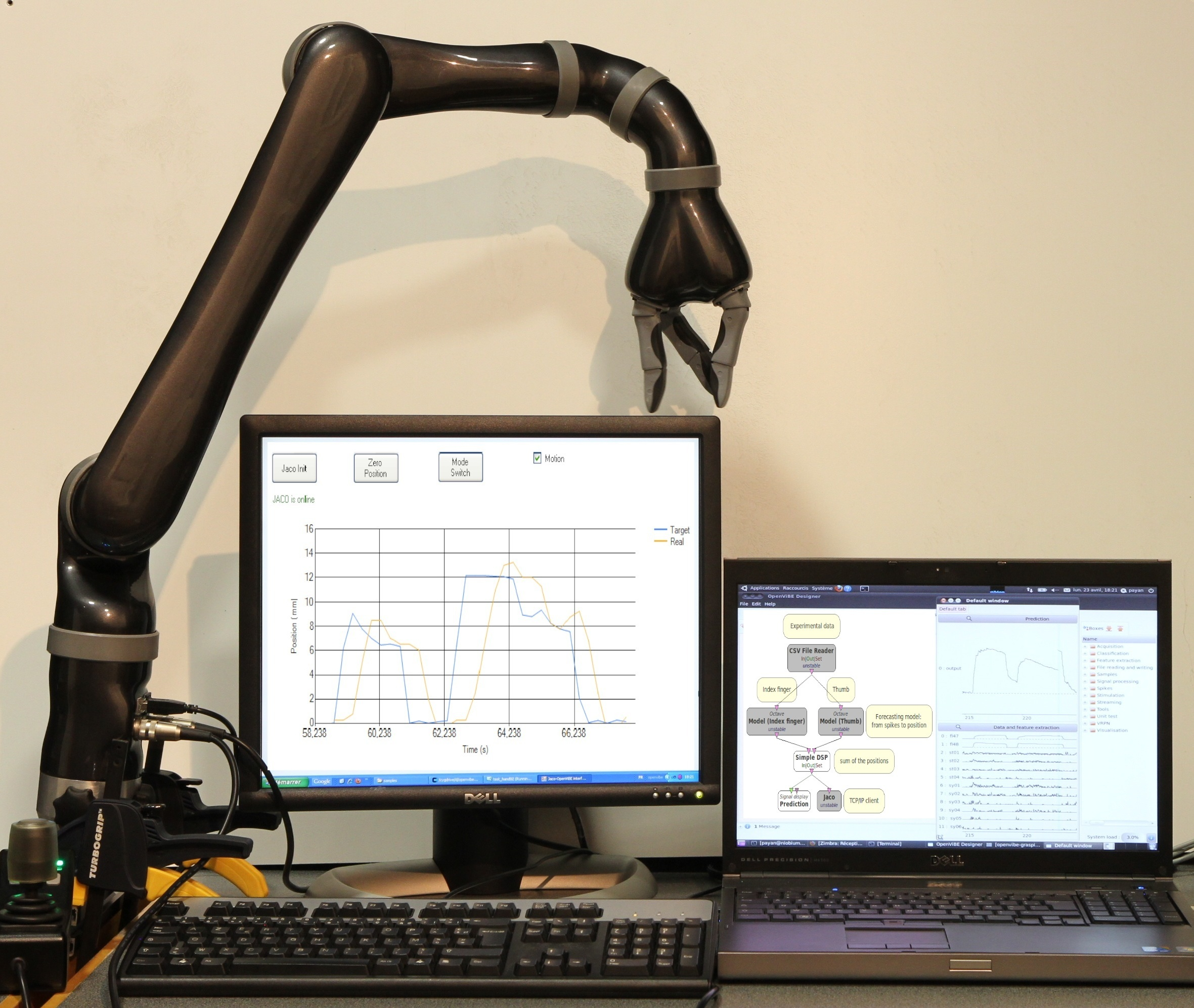}
   \caption{Overview of the experimental setup. OpenViBE (computer on the right side) runs on a linux system to decode intracranial data and send the finger positions to a C\# interface (computer on the left side) that translates them into basic commands for the JACO using the API by Kinova. Screen captures of both computers are respectively available as figures \ref{fig:OpenVibe-Grasp-screen} and \ref{fig:IMG_0523.JPG}.
      }
      \label{fig:Openvibe-JacoCloseUp.png}
    \end{center}
  \end{figure}

We aim at decoding on-line intracranial data corresponding to finger movements recorded in the cortex of a monkey and replicating the movements on a JACO robotic arm by Kinova (Fig. \ref{fig:Openvibe-JacoCloseUp.png}). A specific OpenViBE scenario reads the cortical recordings and predicts the position of the fingers  using a forecasting model.  These positions are sent to an interface that translates the request into basic commands for the JACO arm to execute. Once it has done so, the exact new position is sent back to the interface as feedback and taken into account for the upcoming movements.

 \label{material}
  
\subsection{Dataset}
\begin{figure}[h]
    \begin{center}
     \includegraphics[width=50mm]{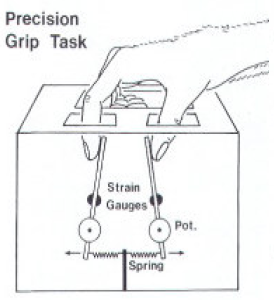}
      \caption{The monkey clasps two independent levers hampered by a spring between  its index finger and thumb (from \cite{Lemon1993}).}
      \label{fig:PrecisionGripTask.png}
    \end{center}
  \end{figure}

We are concerned with reproducing the movement of two fingers. The recordings used in this study have been taken on a monkey (\textit{macaca nemestrina}) that has been trained to perform a precision grip task. The task of the monkey consisted in clasping two independent levers between its index finger and thumb (Fig. \ref{fig:PrecisionGripTask.png}) whenever given a visual GO signal.


\begin{figure}[h]
    \begin{center}
     \includegraphics[width=88mm]{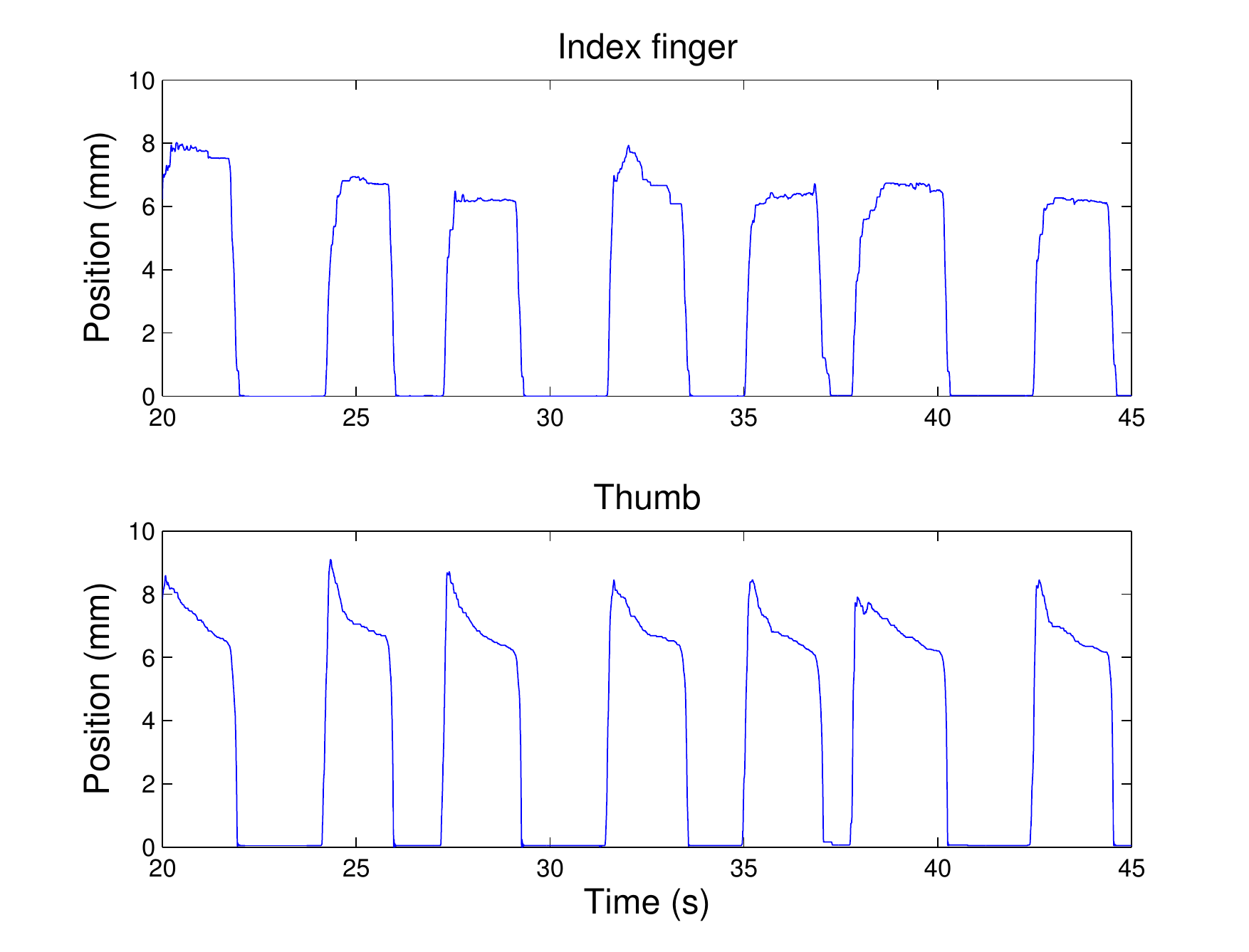}
      \caption{Coordinated finger movements in millimeters, obtained for the index finger (at the top) and thumb (bottom). This is a $25$ seconds excerpt of a $21$ minutes recording.}
      \label{fig:thumb_index}
    \end{center}
  \end{figure}

In these experiments, $33$ corticomotoneuronal cells from the hand area of the motor cortex (area $4$) were recorded with glass-insulated platinum-iridium micro-electrodes (refer to \cite{Lemon1993} for more details about retrieving and filtering the data). Each experiment consists of sequences of neuronal activity signals, or \textit{spike trains}, associated to the recorded positions of the fingers. The muscular effort of each finger is seen as a trajectory over time. See Fig. \ref{fig:thumb_index} for an example of the muscular activities of both fingers. The muscular effort was sampled at $500$ Hz and the neuronal activity at $200$ kHz, the latter has been downsampled at $500$ Hz.
We learn the trajectories of the two fingers separately.

Our model is based on a system of first degree linear equations involving the firing rate of the recorded neurons, a set of thresholds associated to them, some synchrony information and the averaged variation of the levers' positions \cite{Boussaton2012}.

\subsection{JACO robotic arm}
JACO is a robotized arm developed by Kinova\footnote{\url{http://kinovarobotics.com/}}. It is a 6-axis robotic manipulator with a three fingered hand . This upper arm has $6$ degrees of freedom in total. The arm has sensors (force, position and accelerometer) and can reach an object inside a $90$ cm  radius sphere at a maximum speed of $30$ cm/sec. This arm can be placed on a wheelchair and is especially suitable for a person with a disability of the upper limbs. The hand itself has $3$ fingers that can be used in different manners, they can be opened and closed either all together or two fingers: index finger and the thumb. Originally it is possible to use a $3$ axis joystick to control the arm. 

In addition, we were provided with an API by Kinova including basic instructions to interact with the JACO arm.

\subsection{OpenViBE} OpenViBE is a C++ open-source software devoted to the design, test and use of Brain-Computer Interfaces \cite{Renard2010}.
The OpenViBE platform consists of a set of software modules that can be integrated easily and efficiently to
design BCI applications. The key features of the platform are its modularity, its high-performance, its connection with high-end virtual reality displays and it can sustain multiple users. The “designer tool” enables one to build complete scenarii based on existing software modules using a dedicated graphical
language and a simple Graphical User Interface (GUI). This software is available on the INRIA Forge under the terms of the LGPL-V2 license\footnote{\url{http://openvibe.inria.fr}}.

\begin{figure*}[t]
    \begin{center}
     \includegraphics[width=160mm]{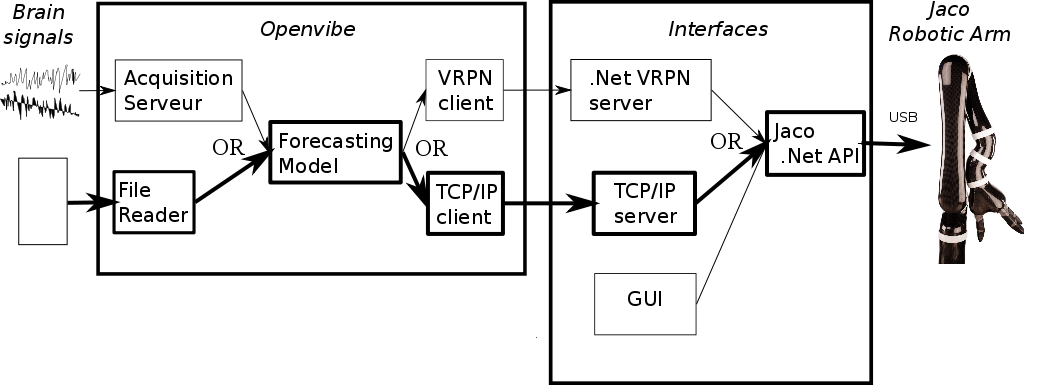}
      \caption{This diagram shows the different parts that we developed to allow the control of a JACO robotic arm. Our contributions can be gathered in two parts: OpenViBE modules and JACO's interfaces. Elements in bold indicate the solution that we finally selected for this experiment.}
      \label{fig:Openvibe2Jacov2.png}
    \end{center}
  \end{figure*}

 \label{method}
Our processing chain allows one to control a JACO robotic arm in a variety of ways (Fig. \ref{fig:Openvibe2Jacov2.png}). First, it is possible to use a GUI that calls the related function of the API by Kinova. Moreover, it is possible to control the robotic arm from several brain signals such as electroencephalographic (EEG) signals or cortical activities. We added specific modules to the openViBE software that have been designed to build the brain machine interface. Then two different client/server applications have been developed to send the finger position and generate the appropriate function of the API to generate an arm motion. We detail our solutions in the following subsections.

\section{Method}

\subsection{OpenViBE modules}
\begin{figure}[h]
    \begin{center}
     \includegraphics[width=88mm]{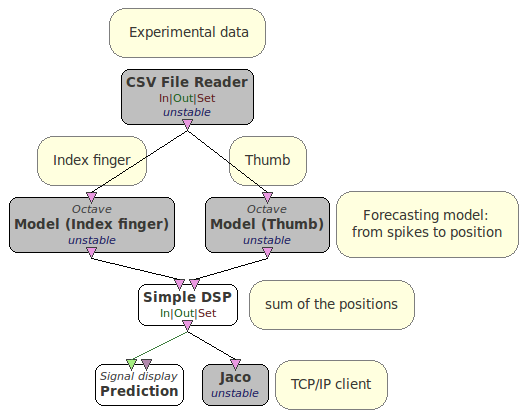}
      \caption{Screen capture of the workflow done within OpenViBE. The OpenViBE designer allows the user to build a scenario through a graphical user interface linking processing modules (grey boxes represent our modules and white boxes the existing modules). The "CSV File Reader" module reads the intracortical data from a datafile and sends them to both the forecasting modules that predict index finger and thumb positions. Then the predicted positions are added together to obtain the distance between fingers. Finally, this value is printed on screen and sent by our client to JACO.}
      \label{fig:OpenVibe-Grasp-screen}
    \end{center}
  \end{figure}

We implemented new OpenViBE modules for building a workflow that read spike trains as inputs and send JACO commands as outputs (Fig. \ref{fig:OpenVibe-Grasp-screen}) 

\subsubsection{Spike modules}

The usual data type in OpenViBE is a set of digitized analog signals,
corresponding to the EEG signals recorded by the BCI devices. The intracranial signals that we consider in this paper are different: they consist of a set of impulses
(or spikes, corresponding to
the action potentials generated by the neurons) that are recorded with a
high timing precision (e.g. 5 $\mu$s).

Fortunately, a common feature in many BCI setups is to use a similar data type for
representing various events (triggers, stimulations, etc.). These
events can be used as a basis for encoding spike trains in OpenViBE, with a 32:32 fixed point precision. We designed a set of new OpenViBE
modules to feed spike trains to the system, and subsequently analyze them,
while making use of already existing modules (e.g. channel selection,
visualization) thanks to the underlying data type used.

\subsubsection{Forecasting model}
The activity of CM cells, located in the primary motor cortex is recorded in the thumb and index fingers area of a monkey. The activity of the fingers is recorded as they press two levers. We defined a  forecasting model that uses a linear combination of the firing rates, some synchrony information between spike trains and averaged variations of the positions of the fingers \cite{Boussaton2012}. This model has been implemented in an OpenViBE module.

\subsection{Interfacing with JACO}
By default, the JACO arm can be controlled using a joystick. Moreover, an API by Kinova was available under windows to read sensors and send commands of movement for a specific direction and a specific duration. This API provides a virtual joystick. This mode of operation does not make it possible to specify the final position of the arm.  Thus, interacting with the JACO arm via the API necessitates the definition of elementary movements.

 \subsubsection{Direct control via a GUI} \label{sec:GUI}
\begin{figure}[h!]
    \begin{center}
     \includegraphics[width=88mm]{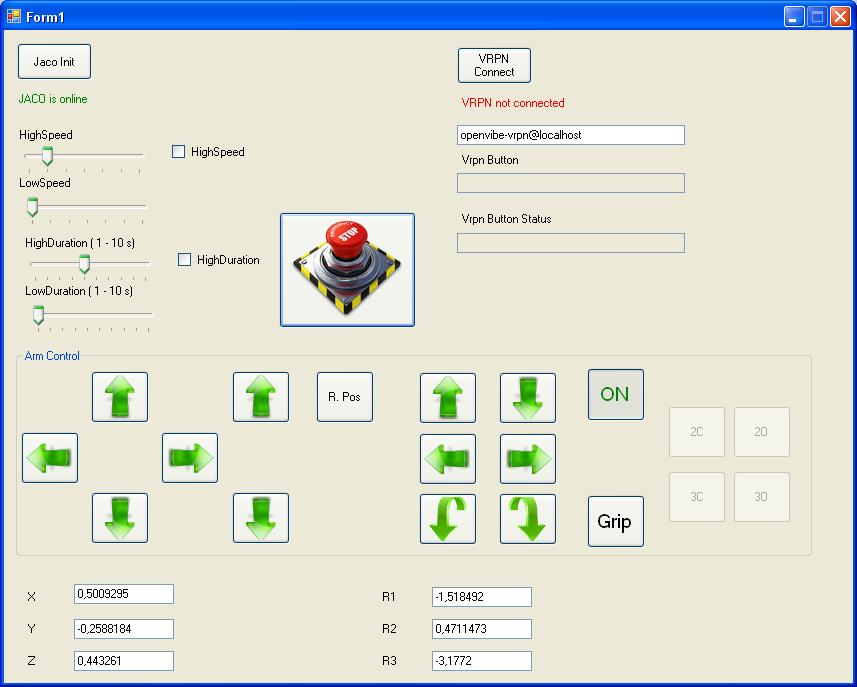}
      \caption{A screen capture of our GUI that allows the user to control the robotic arm via the API by Kinova.}
      \label{fig:interfaceJACO.PNG}
    \end{center}
  \end{figure}

A Graphical User Interface (GUI) for interacting with the arm was designed (Fig. \ref{fig:interfaceJACO.PNG}). The purpose of this GUI was to test the API and produce elementary movements of the arm. 
A sequence of discrete movements then allows the arm to move to the expected position. In order to describe larger movements, the GUI application can record and replay some sequences of tiny movements.
This feature is useful for testing sequences of tiny moves produced by reinforcement learning techniques. 

  The API keeps on improving: it now includes support for Linux platform through Mono\footnote{\url{http://www.mono-project.com}}, it is enriched with new functionalities like angular control and some latencies have been highly reduced.

\subsubsection{VRPN control} \label{sec:VRPN}
The VRPN protocol \cite{Taylor2001} already implemented in OpenViBE is a natural candidate to control the arm. Thus, we built a VRPN client/server using predefined action IDs which can be interpreted by the JACO arm as virtual joystick commands but sent through our application.  
The VRPN server was attached to the PC controlling the JACO arm and a dedicated VRPN client box was created in OpenViBE.
The VRPN server was easily built upon the VRPN .Net package\footnote{\url{http://wwwx.cs.unc.edu/~chrisv/vrpnnet}} under the MIT license. The 1.1.1 release used is although only compatible with .Net 2.0 framework and was not updated recently.  In comparison, the version of the API we used complied with the .Net 3.5 framework on Windows platforms. 
The VPRN server was included in the GUI application defined in paragraph~\ref{sec:GUI}. The recording features also supports the recording of VRPN clients' commands.
This VRPN setup with OpenViBE allows one to control the arm directly with  available applications within OpenViBE like the P300-speller \cite{Farwell88}.
For this purpose, the VRPN interface was designed with 16 commands to control the arm through a 4 by 4 speller. The commands are a subset of the discrete movements provided in the GUI: On/Off, Forward, Forward Left, Forward Right, Backward, BackwardLeft, BackwardRight, Up, Down, Left, Right, Hand / Arm control switch, LowSpeed, HighSpeed, ShortAction, LongAction.

The VRPN communication suffers from a high latency so another client/server based on TCP/IP sockets is now used between OpenViBE and the arm.


\subsubsection{TCP/IP} 
In an attempt to simplify the whole processing chain we replaced the VRPN link
by a custom, simpler network (TCP/IP) connection. This also permitted to
reduce the latency between the OpenViBE framework and the server
controlling the robotic arm, as well as to remove the dependency from
the VRPN libraries.
In addition, instead of a list of dedicated commands that used to
correspond directly to
the atomic controls of the JACO virtual joystick, the OpenViBE pipeline
is now outputting
a target distance for the fingers to follow. This ensure a better
modularity, as we can now hide the robotics details from the BCI side.

\begin{figure}[htbp]
    \begin{center}
     \includegraphics[width=88mm]{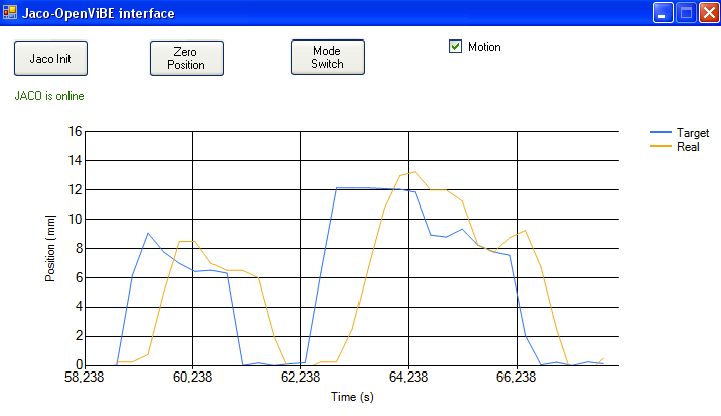}
      \caption{A variation between the requested and the actual position occurs due to transmission delays (sending of the requested position and reading of the sensors value). Here, we can observe a shift of hundreds of milliseconds.}
      \label{fig:IMG_0523.JPG}
    \end{center}
  \end{figure}

This small shift of design also permitted to implement a better control
loop on the robotics
side. The software API that was available at this time\footnote{The newer API
is known to be actually faster} was indeed impaired by a high latency when
using certain USB commands (Fig. \ref{fig:IMG_0523.JPG}). The significant delay we observed between the transmission of motion control and position feedback generated significant oscillations. To improve the closed loop control, we used a conventional solution in control engineering by introducing a proportional control system. This basic solution has been used waiting for the new version of the API that will reduce delays at the origin of the oscillations.

\section{Conclusion}
We built a complete processing chain for interacting with a JACO robotic arm by Kinova.
We illustrated this processing chain with decoding intracranial data recorded in the cortex of a monkey and sending the associated movements to JACO. This chain can be generalized to any movement of the arm or wrist. 

In our daily work, we use OpenViBE to design Brain-Computer Interfaces and we also contribute to its development. Usually, this software interacts with virtual reality applications using the VRPN protocol. So we had to implement a VRPN client/server to send motion commands to JACO via a API by Kinova. We observed a substantial latency between the JACO and openViBE. Thus, we developed another client/server based on TCP/IP sockets that speeded up the whole computation. Here, simplicity is more profitable than powerful but poorly controlled bricks (VRPN). Another difficulty is that the JACO arm has been designed to be controlled using a joystick. This mode of operation forced us to define elementary movements when using the API by Kinova with input data coming from openViBE. Moreover, the usual data type in OpenViBE is a set of digitized analog signals. This format is not appropriate for spike trains. But the modularity of OpenViBE allowed us to easily define our specific format. Finally, when you push software a little outside their comfort zones as we did here using OpenViBE with non standard data and another output protocol, modularity proves its worth. 

In the future, the closed loop control can be improved using the new version of the API which reduces transmission times and using a better PID controller that carries an identification of the dynamics of the fingers. This work will be continued through a realtime control of the JACO using noninvasive EEG signals acquisition systems.

\section*{Acknowledgment}
This work has been supported in part by grants from PRST MISN (R\'egion Lorraine/INRIA).\\ 
This work has been supported in part by CNRS-PIR-Neuro-IC 2010 (Project CorticoRobot).\\ 
The authors are very thankful to R. Lemon and M. Maier for providing the data set to us.\\
The authors would like to thank Baptiste Payan for his expertise and technical support on OpenViBE.

\bibliographystyle{apalike}  
\bibliography{car12}

\end{document}